\documentclass{article}
\usepackage[square,numbers]{natbib}
\usepackage[utf8]{inputenc} 
\usepackage[T1]{fontenc} 
\usepackage{hyperref} 
\usepackage{url} 
\usepackage{booktabs} 
\usepackage{amsfonts} 
\usepackage{nicefrac} 
\usepackage{microtype} 
\usepackage{lipsum}
\usepackage{fancyhdr} 
\usepackage{graphicx} 
\graphicspath{{media/}} 
\usepackage{xcolor}
\usepackage[para]{threeparttable}
\graphicspath{ {images/} }
\providecommand{\keywords}[1]
{
  \small	
  \textbf{\textit{Keywords---}} #1
}

\title{Forecasting Patient Demand at Urgent Care Clinics using Machine Learning

}

\author{
   Paula Maddigan \\
  Massey University \\
  Auckland, New Zealand \\
   \\
  Teo Susnjak \\
  Massey University \\
  Auckland, New Zealand \\
}

\begin{document}
\maketitle
\begin{abstract}
Urgent care clinics and emergency departments around the world periodically suffer from extended wait times beyond patient expectations due to inadequate staffing levels. These delays have been linked with adverse clinical outcomes.  Previous research into forecasting demand this domain has mostly used a collection of statistical techniques, with machine learning approaches only now beginning to emerge in recent literature. The forecasting problem for this domain is difficult and has also been complicated by the COVID-19 pandemic which has introduced an additional complexity to this estimation due to typical demand patterns being disrupted. 

This study explores the ability of machine learning methods to generate accurate patient presentations at two large urgent care clinics located in Auckland, New Zealand. A number of machine learning algorithms were explored in order to determine the most effective technique for this problem domain, with the task of making forecasts of daily patient demand three months in advance. The study also performed an in-depth analysis into the model behaviour in respect to the exploration of which features are most effective at predicting demand and which features are capable of adaptation to the volatility caused by the COVID-19 pandemic lockdowns. 

The results showed that ensemble-based methods delivered the most accurate and consistent solutions on average, generating improvements in the range of 23\%-27\%  over the existing in-house methods for estimating the daily demand. 

\end{abstract}

\keywords{machine learning \and forecasting \and patient demand \and urgent care clinics \and emergency departments}

\section{Introduction}

Urgent Care Clinics (UCCs) and Emergency Departments (EDs) provide continuous urgent medical care to patients requiring assistance. Both UCCs and EDs are frequently the entry-point to the healthcare services for a large segment of patients and are therefore susceptible to periodic overcrowding. Congestion in EDs is particularly problematic since delayed treatment has been linked with numerous negative clinical outcomes \cite{sudarshan2021performance}. However, inadequate staff availability also carries risks which may lead to an unsafe practice resulting from acute conditions being overlooked \cite{Batal2001}.

Therefore it is vital to ensure that strategies are implemented which can efficiently allocate human resources and manage patient demand in these contexts. Effective forecasting of patient demand is one of the strategies which can enable more optimized resourcing of medical services beyond primary care providers. Forecasting of this kind can be in the form of predicting daily patient arrivals as well as more granular predictions at an hourly basis. Forecasting models that also alert in real-time of impending surges in patient arrivals has also been recognized as having high value.

This study develops patient forecasting models for two UCCs providing services to patients experiencing sudden illness or accident-related injury in the Auckland region of New Zealand.  The clinics provide patient walk-in as well as overflow services to local hospitals when their EDs are congested, resulting in ambulances being diverted to them. 

Predicting patient numbers for the UCCs, and similarly for EDs, is a challenging task as patterns from influencing factors have been shown to possess significant variability over time.  The COVID-19 pandemic has added an additional level of complexity from 2020 onward, complicating the existing in-house strategies used by the clinics' administrators for developing effective roster schedules.

Existing literature has primarily focused on predicting patient demand at EDs with limited studies covering UCCs. Generally, most studies have employed traditional statistical methods in preference to machine learning approaches. Some have used Artificial Neural Networks (ANNs) \cite{Khatri2017,yucesan2018multi,khaldi2019forecasting}. Majority of models in literature have incorporated calendar variables and public holiday indicators, with some extending these holidays to the immediately subsequent days and including school holiday flags.  A minority of studies explored features capturing relevant Google search terms \cite{ho2019forecasting}, weather \cite{sudarshan2021performance} and pollution information \cite{sahu2014hierarchical}. However, no literature exists highlighting the use of machine learning to predict patient demand in UCCs within the New Zealand context, having the capability to incorporate mechanisms to handle fluctuating demand caused by the COVID-19 pandemic. 

The aim of this study is to leverage machine learning in order to improve on existing resource planning strategies within the clinics and improve patient-demand forecasting in order to help provide high quality patient care at all times. The study develops models which predict patient demand at a daily basis with the predictions extending three months ahead. The paper presents comprehensive comparisons of forecasting models generated by numerous algorithms and provides analyses into the driving factors influencing patient presentation volumes. The study is specifically targetted for UCCs; however, the methods and findings are also transferable for EDs.  \par

The research questions (RQ) addressed in this paper are:
\begin{itemize}
\item (RQ1) Are machine learning models able to improve on existing in-house strategies for forecasting patient demand at UCCs?

\item (RQ2) What are the most effective machine learning algorithms for predicting daily demand at UCCs across a three-month forecasting horizon?

\item (RQ3) Which features are the key drivers in predicting patient demand at UCCs?
\end{itemize}

\section{Literature Review}

Given the paucity of prior research in forecasting demand for UCCs as well as some similarity between UCCs and the EDs, we include in this literature review studies that have focused on EDs. While UCCs are specifically designed to manage lower-acuity conditions than EDs, it has been estimated that up to around a quarter of all ED visits could effectively be serviced by UCCs \cite{carlson2020impact}. Findings\footnote{Royal New Zealand College of Urgent Care}  indicate that cities with UCCs have significantly lower ED presentations, thus confirming the correlation between the patient demand for EDs and UCCs, and ultimately the existence of similar drivers of demand across both types of medical facilities. Therefore, approaches that have been effective on the ED domain include transferable insights for the UCC context, and vice versa. Indeed some prior studies \cite{Wargon395} have used literature from EDs and UCCs interchangeably.     

Some of the earliest forecasting research into patient demand has made use of the simple modelling approaches like linear regression for time-series prediction by making adjustments to the input data to incorporate calendar information. Seasonal and cyclic effects can be added using dummy variables representing the day of the week, month of the year, public and school holiday flags and more. \citet{Batal2001} successfully used this modelling approach at the UCC located at the Denver Health Medical Center in Colorado, USA.  With the inclusion of weather and calendar variables to the model, they showed weather contributed little value to the accuracy of predictions.  However with incorporating a selection of calendar variables like day of the week, month, season, and holiday flags, they believed their regression models produced more accurate forecasts of patient numbers. The average daily patient volume at the clinic varied between 115 on Mondays decreasing steadily down to 69 on Sundays and their study highlighted the expected increase in patients during the winter months. Interestingly they discovered an 11\% increase in patient presentations following a holiday, with patients choosing to delay medical help to avoid holiday disruption or participating in holiday activities, thus seeking medical help the next day.

\citet{Jones2008} used a similar approach with multiple linear regression to create a benchmark model for predicting patient presentations in three hospital EDs in Utah and Idaho, USA.  They included calendar variables representing day of the week, month of the year, and holiday flags. Unlike the clinics studied by \citet{Batal2001}, patient presentations were higher in the weekend and dipped mid-week. They enhanced their model to include interaction terms between calendar variables and climatic variables.  By accounting for serial autocorrelation using ARIMA models, they saw little improvement in the accuracy of their predictions.  The ARIMA models were enhanced to SARIMA models to account for seasonality, however they noted that these univariate models could not incorporate other factors like calendar and weather variables.  After considering an exponential smoothing model and a neural network model, they concluded their regression models with calendar variables resulted in the best predictions, with potential improvement in accuracy by including location specific information such as special days.  Their research used a forecast horizon of 30 days with an iterative testing strategy. The model was estimated on a training set and measured on an unseen 30 day test set. 

\citet{Aboagye_Sarfo_2015} used both multivariate vector VARMA and univariate ARMA time series models to forecast patient demand in public hospital emergency departments throughout the state of Western Australia.  Their work identified the VARMA multivariate models were more accurate than univariate ARMA models to predict demand for the purpose of resource allocation and strategic planning.  The study did not incorporate drivers for patient presentations, however it identified seasonal trends by age group, place of treatment, triage category and disposition. The data was supplied on a monthly granularity, limiting its usefulness. However the models were built on 5 years of data and tested using a 2 year period, with MAE, RMSE and MAPE  evaluation metrics calculated, providing a more stable evaluation than \citet{Jones2008}. 

Part of the study carried out by \citet{Boyle2011} analysed ED presentations in two Queensland hospitals in Australia over 5 years from 2002 to 2007.  They used multiple regression, ARIMA and exponential smoothing models to predict daily demand, evaluating performance with the MAPE metric only.  They noted that days with matching characteristics, like day of week and public holidays tended to have the most similar patient demand. Highlighting the importance of testing model performance over both the summer and winter months, they allocated the final year in the dataset  to the testing period. Rather than using a gradually expanding training set and a moving test set, they kept the test set static and used 4 training sets, from 1 to 4 years in duration. To investigate the applicability of their models to other emergency departments, they tested them on data from 2005 to 2009 for 27 other Queensland hospitals.

\citet{Champion_2007} built forecasting models to predict the monthly demand at a hospital in the Australian state of Victoria.  Using data from 2000 to 2005, they used exponential smoothing and ARIMA methods to predict demand for the first 5 months of 2006.  Using RMSE metrics to compare models, they concluded their best exponential smoothing model with a simple seasonal component had smaller error than their best ARIMA(0,1,1) non-seasonal moving average model. However resulting forecasts from both models were similar.  Their exponential model predicted seasonal fluctuations with ease however no performance improvement was seen when including seasonal terms into the ARIMA model.  But as previously discussed, measuring model performance with less than a 1-year cycle of data may produce misleading conclusions about forecast accuracy, and forecasting at such large granularity of monthly demand provides limited usefulness in resource allocation.

\citet{Jilani2019} present a unique study using Fuzzy Time Series (FTS), a concept taken from Fuzzy Logic \cite{Aladag2017}.  Using data from 2011 - 2015 over four emergency departments in the UK, they predicted daily demand for a 4 week horizon. They created a time-series model for each day of the week as a method of modelling week day variations, then built a FTS model for each day.  Performance was measured using MAPE and RMSE.  Comparing their FTS models performance to that of ARIMA and neural network models, they showed their model was twice as accurate as the ARIMA models.\par

The most similar approach in literature to the work presented in this study is that of \citet{Vollmer_2021}. They discuss the extra standard the NHS in the UK introduced in 2010, with the requirement that 95\% of emergency department patients being seen within 4 hours of arrival. This target was only met until 2014 which \citet{Vollmer_2021} believe is due to inefficient staffing.  Their analysis used 8 years of daily data from St Mary’s and Charing Cross hospitals, and applied traditional time series as well as machine learning algorithms to predict patient demand 1, 3 and 7 days ahead.  They incorporated features such as seasonal patterns, weather, school and public holidays as well as large scheduled events which tended to see an influx of patients.  They also utilised google search data for the keyword “flu”.  Their models included 4 traditional time series models, including ARIMA and exponential smoothing, and 4 machine learning algorithms – an extension to the generalised linear model, a type of Random Forest, a Gradient Boosted Model and k-Nearest Neighbours.  Even with applying cross-validation techniques and hyper-parameter tuning to the machine learning models, they concluded the quality of forecasts from traditional linear models were comparable to machine learning models using the MAE metric. Utilising the power of ensemble methods, they added a Stacked model using both their time series and machine learning algorithms to deploy a more accurate and robust model. Interpretation of their models was only taken as far as studying variable importance by way of permutation importance, calculated as the decrease in the error when the feature is randomly shuffled \cite{Breiman_2001}.

\citet{Duarte_2019} took a difference approach. Rather than forecasting demand, they forecasted key performance indicators, with one being the average number of minutes patients waited in the emergency department before being seen.  Data was aggregated on an hourly basis from Medway Foundation Trust's Emergency Acute Unit in Kent, South East England. They built an ARIMA model as a benchmark, and showed Facebook's Prophet model improved forecast accuracy. The work of \citet{Duarte_2021} extended this analysis with one of the KPI's being the average number of patients waiting in the emergency department each hour.  As well as ARIMA and Prophet models, they included a General Regression Neural Network (GRNN), including data throughout the first COVID-19 lockdown and beyond.  They observed ARIMA and Prophet taking 5 weeks to react to the change in demand from the COVID-19 outbreak; however, GRNN adjusted quickly to the change.  Interestingly they noted predictions improved for all models after lockdown as the variance of the forecasts were diminished.

A number of other studies have been undertaken with the aim of identifying peak demand days. 
\citet{Khatri2017} used machine learning to predict these outcomes for chronic respiratory illnesses presenting at emergency departments in Dallas County using Neural Networks and Random Forest. They incorporated weather and pollution data into the models, but concluded with the large number of contributing factors to patient admissions, forecast accuracy was limited by the available data.

\section{Methodology}
\subsection{Experimental Design}

\paragraph{Setting} The data was acquired from two clinics, owned by Shorecare\footnote{https://www.shorecare.co.nz/}. The Smales Farm clinic offers 24-hour care, while the Northcross clinic offers after-hours care. The clinics are equipped to manage a range of medical problems as well as x-ray and fracture clinics, and facilities for complex wound management. The Smales Farm clinic is the only 24-hour UCC within a catchment area of a population of approximately a quarter of a million and is located within a kilometer of a major hospital whose ED treats $\sim$46,000 patients annually.

\begin{figure}[htb]
\centering
\includegraphics[scale=0.5]{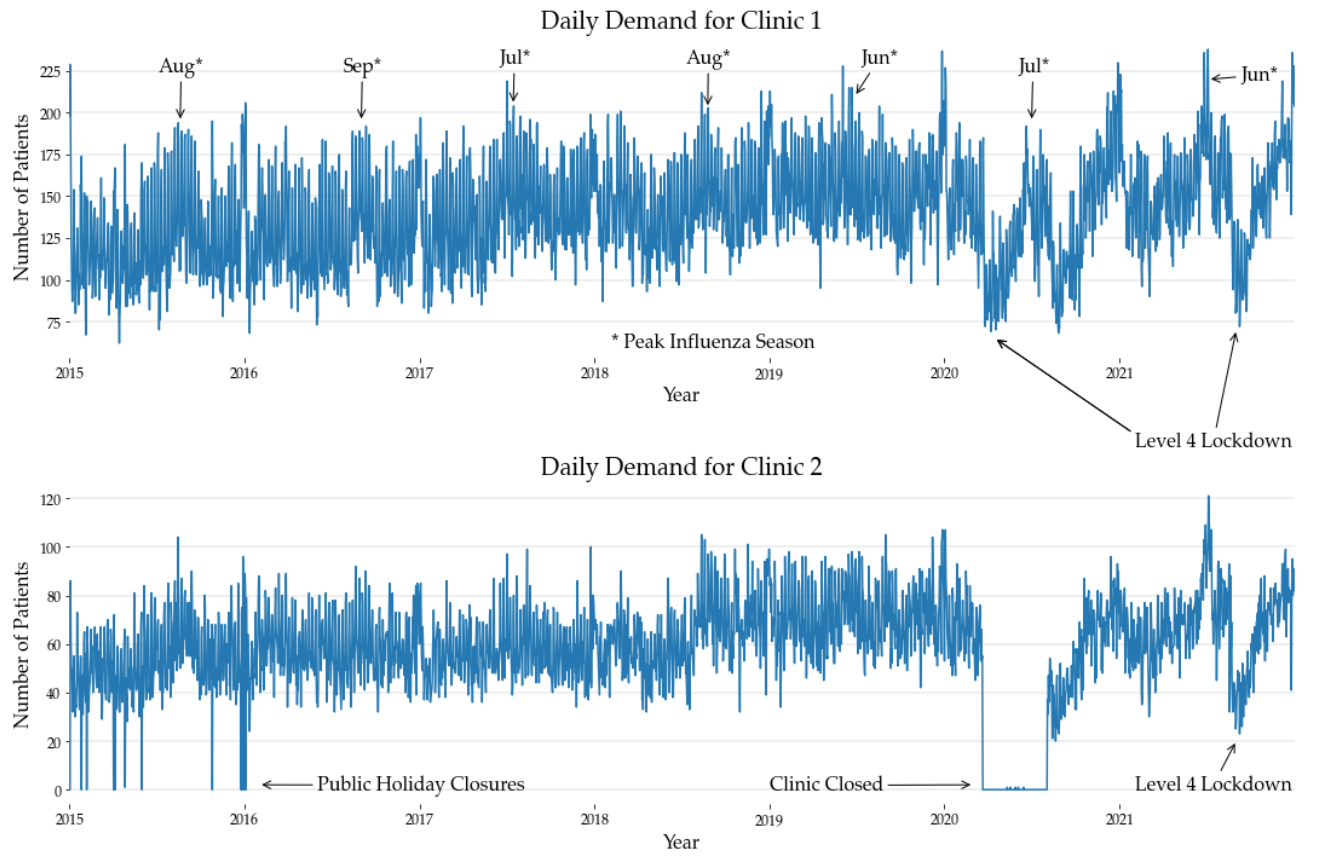}
\caption{Daily Demand}
\label{figure:ts}
\end{figure}

\paragraph{Dataset} Models were designed for predicting daily demand three months (13 weeks) in advance. Data was provided from 2011 through to 2021, and aggregated to a daily granularity. Figure \ref{figure:ts} depicts the characteristics of a subset of the data for both clinics.

\paragraph{Benchmark models} In order to robustly evaluate the efficacy of the candidate models, we created several benchmark models for comparisons. The first benchmark model approximated the current in-house approach used by the clinics to estimate demand. This strategy involved adding a 5\% increase to the patient numbers from the same period in the previous year. The second benchmark model we refer to as the Na{\"i}ve model, represented a Random Walk method, taking the forecasted value to be the same as the value for the same period of the previous year. The third benchmark model was generated using ARIMA. An additional enhanced version of the Na{\"i}ve model was also developed which attempted to generate forecasts for a given day at a point in time \textit{t} by calculating a mean value of weighted lags \textit{t}-7, \textit{t}-14,  \textit{t}-364, \textit{t}-728 and \textit{t}-1092. 

\paragraph{Algorithms} We used 12 statistical and machine learning algorithms in order to generate competing models. These included: Random Forest (RF), Voting, Stacking, Ridge Regression, K-Nearest Neighbour as implemented in Scikit-learn \cite{scikit-learn}, CatBoost \cite{catboost}, Prophet \cite{taylor2018forecasting}, an averaging model (5 base models, discarding the highest and lowest predictions) and four models with  an error correction component (this involved adjusting predictions using exponential smoothing as well as autoregressive approaches to forecast residual error).

\paragraph{Features} With strong weekly, yearly and seasonal patterns, feature variables were added representing lags of 7, 14, 364, 728 and 1092 days.  Adding the week of the year calendar variable accounted for the increasing trend throughout the year and seasonal trends including school holiday patterns. A public holiday flag ensured the elevated demand on these days was represented.  To accommodate for COVID-19, an indicator was included representing the legally mandated restrictions in place within the region of the clinics, either through the COVID-19 Alert Level system \cite{NZG2022} or the Traffic Light mappings \cite{NZG2022a} defined by the Government. Table \ref{table:FeatureDesc} lists the names of all the variables used and their descriptions.

\begin{table}[h!]
\centering
\begin{tabular}{l  l}
\hline 
Feature & Description
\\
\hline 
lag7d  & 7 Day Lag,  \textit{t}-7 value, one week prior\\
lag14d & 14 Day Lag,  \textit{t}-14 value, two weeks prior \\
lag1 & 364 Day Lag,  \textit{t}-364 value, one year prior \\
lag2 & 728 Day Lag,  \textit{t}-728 value, two years prior \\
lag3 & 1092 Day Lag,  \textit{t}-1092 value, three years prior \\
public\_holiday & Public Holiday Indicator (0/1)\\
week & Week Number ranging from 1 - 53\\
covid\_level & COVID-19 Alert Level ranging from 1-4,\\ 
 & or Traffic Light (Green=1; Orange/Red=2)\\
     \hline
\end{tabular}
\caption{Feature Descriptions}    
\label{table:FeatureDesc}
\end{table}

\paragraph{Testing approach}
Models were tested on data over 3 years from 2017. An expanding window approach was used for testing the models. The models were initially trained on data from 2014-2016 and the forecasts were made from 1 January 2017 up to 13 weeks ahead (one quarter). For each quarter, the models initially forecasted demand one week ahead, with all lag variables available in the training set. In predicting week 2, the forecasted demand values for the week prior were used in the 7 day lag variables, with actual values from the training set used for 14 day lag values.  Subsequent weeks required forecasted demand values to be used in both the 7 day and 14 day lag variables. This meant that the input variables for subsequent predictions were prior estimates generated by the model.

Following the prediction of the values for the upcoming quarter, the training window would then be expanded to include the actuals from the next quarter and the forecast horizon would then shift and the demand would be predicted for the subsequent quarter. By continuously repeating the forecasting process until the end of 2019, 12 forecasts periods were generated, each containing daily forecasts for a 13 week period. 

Forecasts were not made for the 2020 data due to the disruption caused by the pandemic and lockdown mandates which resulted in the forced closures of the clinics during extended periods of time during this year. 
Given these COVID-19 disruptions, we present and analyse the model forecasts separately for years 2017-2019 and for 2021.

\paragraph{Model evaluation} Algorithms were ranked according to RMSE performance within each forecast period (quarter), with averages calculated over the 12 quarters. The calculation of the mean ranks for each algorithm across each quarter were calculated and used for producing effective comparisons of all models. Mean Absolute Error (MAE) values and Mean Absolute Percentage Error (MAPE) values were also calculated and presented. Consideration was given to enhancing the model performance and re-running the testing methodology by implementing hyper-parameter tuning and 5-fold cross validation.  
The \citet{Diebold1995} test was used on forecast results to establish if model predictions were significantly different from benchmark models.

\paragraph{Model interpretability} We evaluate the behavioural mechanics of the predictive models at both global and local levels. At a global level, we consider the overall aggregate effects that each feature has on the model outputs. For this, we rely on feature importance plots which rank as well as depict relative impacts of each feature. We also consider feature importance visualisations which have the ability to depict the effect of changing feature values on the final forecast. These plots together offer a degree of high-level interpretability of the key drivers for a given model.

When we consider model behaviour at a local level, we are seeking an explanation from a model in terms of how exactly it has arrived at a given forecast for a specific data point. For this, we use both  LIME \cite{Ribeiro_2016} and SHAP \cite{Lundberg2017}. Both tools provide a valuable insight into which features and their values are most influential in increasing or decreasing overall forecasted demand.

\section{Results}
\subsection{Performance comparisons}

\begin{table}[tb]
\centering
\begin{threeparttable}
\begin{tabular}{l c c c c c c c}
    \hline
 & \multicolumn{3}{c}{Clinic 1} &   
      \multicolumn{3}{c}{Clinic 2}   & \\
    Model & RMSE & (MAE) & R\tnote{*} & RMSE & (MAE) & R\tnote{*} & R\tnote{+}\\
    \hline
     Current &  21.9 & (17.7)  &  13.8 & 13.6& (10.9) & 12.8 & 13.3\\
     Na{\"i}ve &  22.9 &(18.5)  & 14.2  & 13.8 &(11.1) & 12.6&  13.4 \\
     ARIMA &  23.8 &(20.0) & 10.7  & 11.8 &(9.7) & 9.1& 9.9\\
     Na{\"i}ve Enhanced &  20.4 &(16.6)  &  11.4 &  11.4 &(9.3) &6.8 &  9.1\\
     Random Forest & 17.6& (13.9)  &  8.1 & 11.7 &(9.1) &7.9&8.0\\
     CatBoost & 18.0 &(14.4) & 8.6 & 11.8   &(9.3) & 9.0& 8.8 \\
     \textbf{Prophet} &  \textbf{16.4}& \textbf{(13.3)} & \textbf{4.9} &  \textbf{10.9}  & \textbf{(8.2)} & \textbf{6.8}& \textbf{5.9} \\
     \textbf{Voting} &  \textbf{15.9} &\textbf{(12.8)} & \textbf{4.7} &   \textbf{10.6}  & \textbf{(8.2)} & \textbf{5.3}& \textbf{5.0} \\
     \textbf{Stacking} &  \textbf{16.2} &\textbf{(13.1)}  & \textbf{4.6} &   \textbf{10.3} &  \textbf{(8.0)}  & \textbf{4.1}&\textbf{4.4} \\
     Ridge &  17.3 &(13.9) & 8.1 &   11.4  & (8.8) & 7.3& 7.7 \\
     KNN &  19.3 &(15.3) & 11.0 &   11.5  & (9.2)  & 7.8& 9.4\\
     \textbf{Averaging} &  \textbf{16.0} &\textbf{(12.9)}   & \textbf{5.3} &  \textbf{10.6} &\textbf{(8.2)}  & \textbf{5.1} & \textbf{5.2}\\
     CatBoost AutoReg &  18.1 &  (14.5) & 7.9 &  11.9& (9.4) & 10.1& 9.0\\
     CatBoost Smoothing &  19.2 &(15.4)  &  9.3 &  12.5 &(9.9)  & 10.5& 9.9\\
     Prophet AutoReg &  17.8& (14.6)  & 5.8 &  11.9 &(9.4)  &10.3 & 8.1\\
     Prophet Smoothing &  19.0 &(15.5) & 7.5 &  12.2& (9.8)  &10.8 & 9.2\\

     \hline
\end{tabular}
\begin{tablenotes}
 \item [*] Mean ranking per clinic
 \item [+] Mean ranking over both clinics 
 \end{tablenotes}
\end{threeparttable}
\caption{Model Metrics Forecasting 2017-2019}    
\label{table:AvMet}
\end{table}

For each model, Table \ref{table:AvMet} shows the average RMSE and MAE with the mean rankings for the 12 forecasts generated from the iterative prediction method between 2017 and 2019. The top four performing models with the lowest mean rankings (best ranked) and metrics up until 2019 for Clinic 1 (Smales Farm) are Voting (15.9), Averaging (16.0), Stacking (16.2) and Prophet (16.4). The patterns were similar for Clinic 2 (Northcross), with Stacking generating the lowest error. The results indicate that the efforts to implement error correction mechanisms for adjusted model predictions of CatBoost and Prophet forecasts using AutoRegressive and Smoothing models did not yield improvements over the base models alone.

While Table \ref{table:AvMet} indicates that the generated models perform more accurately than the benchmark modes, we performed additional analyses in order to determine if the forecasts from the best performing algorithms are indeed significantly different at a statistical level, from the forecasts of the current predictive models used at the clinics. For this, we relied on the Diebold-Mariano test. Our results indicated that each model pair calculated from the concatenation of the 12 forecasts generated between 2017 and 2019 is significantly different (at the  0.01 level) to the benchmark forecasts. In terms of assessing pair-wise differences between the forecasts of the best performing models, we found that for Clinic 1, most model forecasts are significantly different (at the 0.05 level) except Stacking, Averaging and Prophet.  For Clinic 2 forecasts, most forecast pairings can be considered significantly different to each other, except CatBoost/Random Forest, Voting/Prophet, and Averaging compared with Stacking, Voting or Prophet.

\begin{table}[htb]
\centering
\begin{threeparttable}
\begin{tabular}{l  c c c c c  c c c c c c}
\hline
 & \multicolumn{5}{c}{Clinic 1} &
      \multicolumn{5}{c}{Clinic 2} \\
     & Q1& Q2 & Q3 & Q4 & R\tnote{*} & Q1 & Q2 & Q3 & Q4 & R\tnote{*}& R\tnote{+}\\
    \hline
    RF & 21.6  & 21.0  & 19.9  & 46.5 & 4.3 & 21.3 & 14.9  & 16.8  & 26.8 & 2.8 & 3.5\\
    CatBoost & 20.9  & 22.7  & 23.4  & 49.3  & 5.3 & 30.9  & 12.4  & 14.5  & 28.7 & 3.5 & 4.4\\
    Prophet & 18.6  & 20.2  & 20.4  & 32.8 & 2.5 & 24.0  & 22.6  & 21.3  & 17.2& 3.5 & 3.0\\
    \textbf{Voting} & \textbf{17.1}  & \textbf{19.7} & \textbf{18.7}  & \textbf{39.9}  & \textbf{1.3} & \textbf{26.6}  & \textbf{17.1}  & \textbf{13.9}  & \textbf{22.9} & \textbf{3.0} & \textbf{2.1}\\
    Stacking & 20.6 & 26.2  & 22.3  & 46.3 & 4.8  & 36.6  & 21.7  &21.6  & 50.2 & 5.8 & 5.3\\
    Averaging & 18.4  & 23.0  & 19.4  & 41.7  & 3.0 & 26.4  & 15.5  & 13.1  & 25.1 & 2.5 & 2.8\\
     \hline
\end{tabular}
\begin{tablenotes}
 \item [*] Mean ranking per clinic
 \item [+] Mean ranking over both clinics 
 \end{tablenotes}
\end{threeparttable}
\caption{Model RMSE Forecasting 2021}    
\label{table:2021Met}
\end{table}

The best performing models on the 2017-2019 data from Table \ref{table:AvMet} were analysed for their forecasting accuracy on the 2021 data. This can be seen in Table \ref{table:2021Met}.   Mean rankings show that for these testing data, Voting is superior for Clinic 1, ranking 3rd for Clinic 2. However, over both clinics, Voting is still the best performing algorithm on average on these datasets.

\begin{table}[htb]
\centering
\begin{tabular}{l c c c c}
 \hline
 & \multicolumn{2}{c}{Clinic 1} &
      \multicolumn{2}{c}{Clinic 2} \\
    Model & 2017-2019 & 2021 & 2017-2019 & 2021\\
    \hline
     Random Forest & 9.6 & 14.8  &  14.1 & 24.0 \\
     CatBoost & 9.9 & 16.0 &  14.3  & 28.7 \\
     Prophet &  9.3  & 12.1  & 13.2  & 28.0 \\
     \textbf{Voting} &  \textbf{8.9}  & \textbf{12.7}  &  \textbf{12.8}  & \textbf{25.7} \\
     Stacking &  9.1  & 15.7  &  12.8  & 44.8 \\
     Averaging &  9.0  & 13.8  &  12.8  & 25.9 \\
     \hline
\end{tabular}
\caption{Model MAPE Forecasting 2017-2019 and 2021}    
\label{table:MAPE}
\end{table}

Table \ref{table:MAPE} highlights the overall differences in the predictability of demand across both clinics using the best performing algorithms identified above. The results are portrayed as MAPE to achieve normalisation which is necessary for making the predictive results comparable between the clinics. Values are higher for Clinic 2, indicating the models generally do not perform as accurately for this clinic. The predictability of Clinic 2 can be seen to deteriorate especially on the 2021 data in comparison to Clinic 1. We can infer from this that Clinic 2 has been more susceptible to pandemic disruptions than Clinic 1 and the volatility in the actual data that supports this claim can be seen in Figure \ref{figure:ts}.

Table \ref{table:Improve} lists the average percentage improvements in MAE for each top performing model over the benchmark in-house model used by the clinics for the 2017-2019 test period. Voting provides the largest improvement over the current model for Clinic 1, with an average increase in accuracy of 27\% on forecasting a 13 week period. Voting performance improvement for Clinic 2 is only slightly inferior to Stacking and Averaging.\par

\begin{table}[htb]
\centering
\begin{tabular}{l c c c c}
    \hline
    Model & Clinic 1  & Clinic 2\\
    \hline
     Random Forest & 21\%  &  16\%  \\
     CatBoost & 18\%  &  14\%   \\
     Prophet &  24\%   &  21\%   \\
     \textbf{Voting} &  \textbf{27}\%   &  \textbf{23}\%   \\
     Stacking &  25\%   &  24\%   \\
     Averaging &  26\%  &  24\%   \\
     \hline
\end{tabular}
\caption{Model Improvements over current in-house benchmark model for data ranging 2017-2019}    
\label{table:Improve}
\end{table}

\subsection{Forecast Stability}

\begin{figure}[htb]
\centering
\includegraphics[scale=0.46]{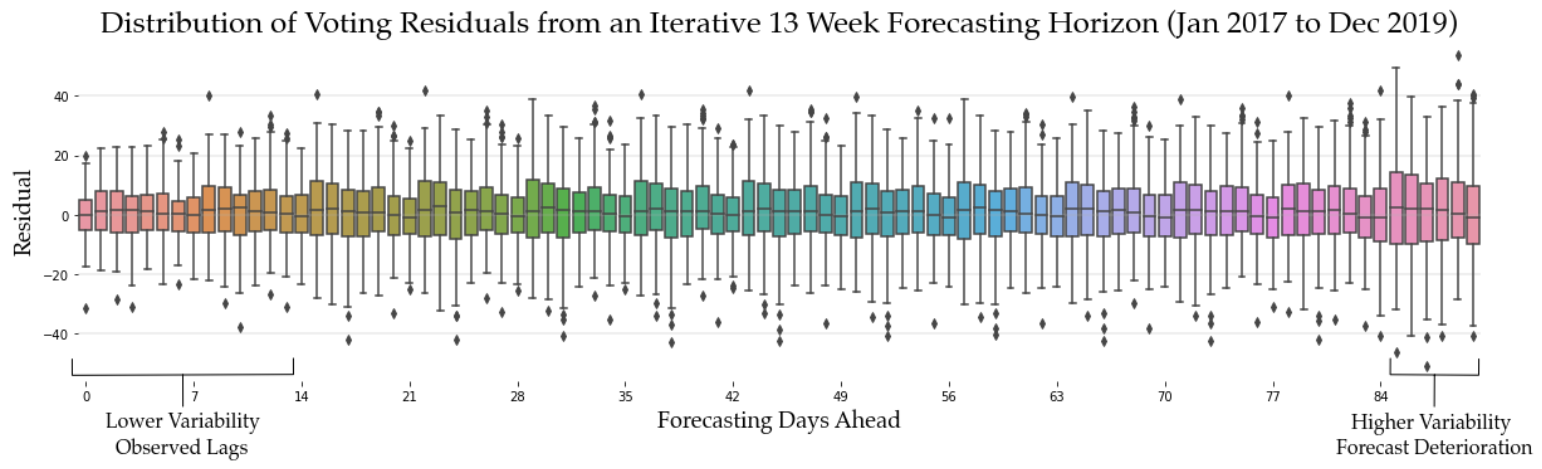}
\caption{Forecast Stability for 13 Week Prediction Horizon for Clinic 1}
\label{figure:stab}
\end{figure}

Figure \ref{figure:stab} shows the variability in the forecast errors of the Voting model over the 13 week horizon for Clinic 1, from Jan 2017 for 3 years. The figure represents the aggregate variability of the forecasts across all the forecast vintages that were tested. The figure highlights the presence and the degree of deterioration in forecast accuracy as the forecasts extend towards the maximum extent of 13 weeks. The figure indicates that the model reliability is relatively constant regardless of the projected prediction date.  Variability in the first two weeks is smaller due to the forecasts utilising actual one week lag data in the first week and observed two week lag data in the first fortnight instead of forecasted values. Around the 85 day mark, some increase in variability can be detected which may indicate that the forecast deterioration begins to take place at this point.

\subsection{Model Interpretability}

We conduct a deeper analysis of the Voting model since this is the best performing model on average across both clinics. We use SHAP and LIME tools in order to extract model interpretability and forecast explainability insights. Additionally, we provide feature importance graphs from the perspective of CatBoost and Random Forest models separately since they are multivariate algorithms and are able to provide sufficient indication on the behaviour of the overall Voting model in respect to effects of different features on the model outputs. These models are used also since the Voting model is constructed from three base estimators - CatBoost, Random Forest and Prophet - with Prophet being a univariate algorithm.

We highlight several diverse forecasting vintages as examples of the model behaviour. For this, we use the Voting model's behaviour on forecasting values for the 2018 Quarter 2 and 2019 Quarter 4, which represent relatively stable data patterns. We contrast the model behaviour from these stable periods with the model behaviour for forecasting 2021 Quarter 1, which encompassed a highly disruptive period due to pandemic lockdowns. We then focus specifically on examining the model behaviour on forecasting demand for atypical scenarios such as holidays which are known to strongly affect patient presentation volumes - for this we analyse model explainablity for forecasting Easter and school holiday patient demand, while examining the magnitude of impact of the various features.

\subsubsection{2018 Quarter 2 forecast horizon}
\begin{figure}[htb]
\centering
\includegraphics[scale=0.42]{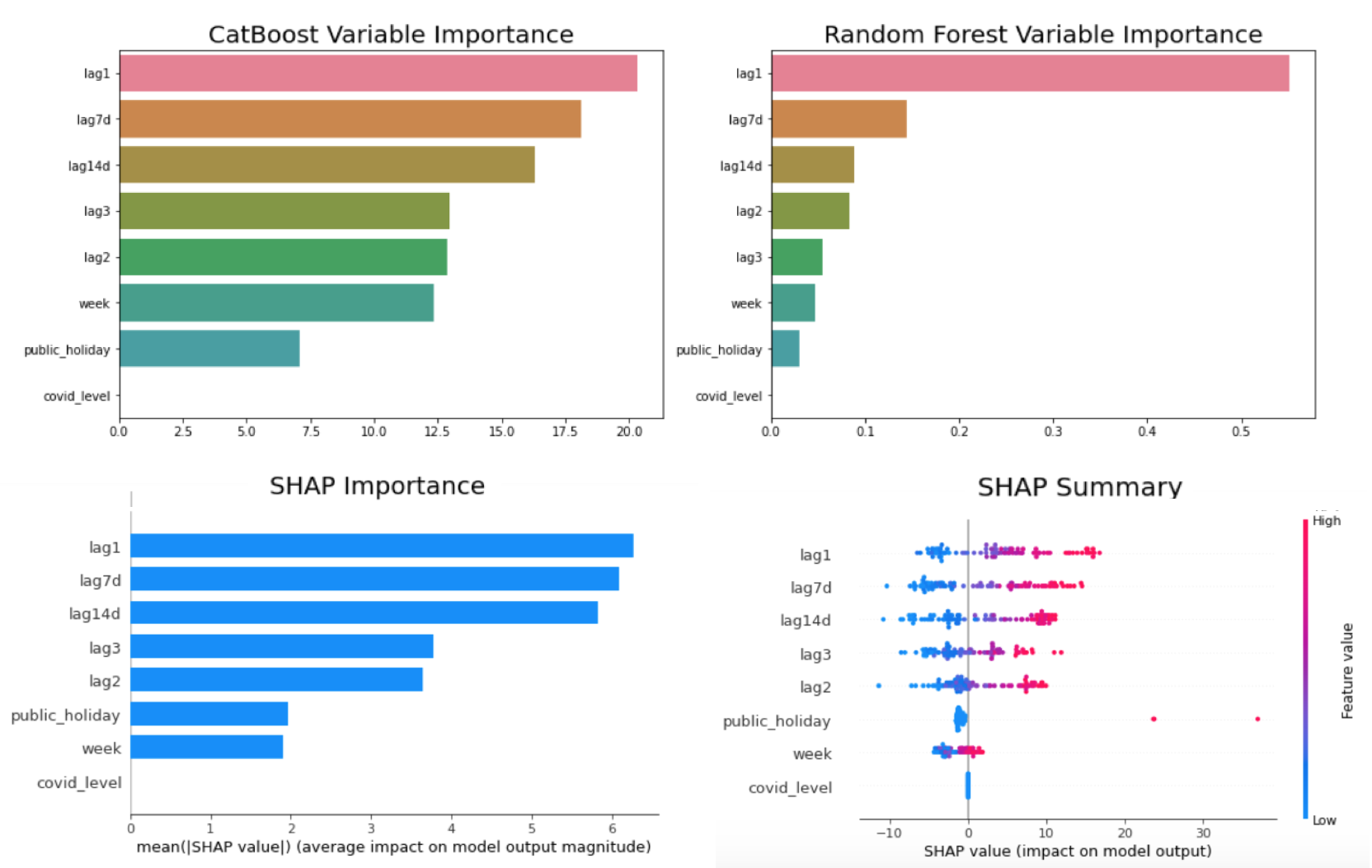}
\caption{Variable Importance Plots Q2 2018}
\label{figure:imp_2018}
\end{figure}

The variable importance plots in Figure \ref{figure:imp_2018} generated from the forecast in Quarter 2 of 2018 show the most important feature influencing patient presentations is the demand as recorded from the same day one year prior. The demand exactly one week prior is the next important, followed by the demand a fortnight before. The Random Forest model places heavy weighting on the 1 year lag feature compared to CatBoost, with CatBoost tending to agree more with the generated SHAP importances.\par

The SHAP Summary plot, like the variable importance plots, arrange features by decreasing importance, with the most important feature at the top. The SHAP value for each data point is plotted for each feature, with red denoting the feature has a high value, blue a low value and purple near the average value for the feature. The x-axis represents the SHAP value itself.  Points with a positive SHAP value (appearing to the right of the vertical zero line) have a positive impact on the forecast value, hence they contribute towards predicting a higher demand. Those points having a negative SHAP value (to the left of the vertical zero line) have a negative impact and decrease the forecasted demand. The further the points extend from zero, the larger the contribution to the final prediction.  \par

The SHAP Summary plot in Figure \ref{figure:imp_2018} generally shows that all lag values are red to the right of the zero line and blue to the left.  Hence as we observe overall higher lag values, this corresponds to a higher  forecasted demand. The reverse also holds. Purple values around the mean have small positive SHAP values and hence marginally contribute to increasing the forecasted demand. The week number generally causes a decrease in demand, but as the year progresses and the week number increases, demand sees less of a reduction in its value, until approximately the last quarter of the year when the week number increases demand. Three days in this period are flagged as public holidays which have a very large influence on pushing up the predicted value (Easter Monday and Queen's Birthday with SHAP values of 24, ANZAC Day with a value of 37).  Non-public holiday days result in a reduction of forecasted demand. The detailed influence of various holidays on forecasted predictions will be presented separately. 

\subsubsection{2019 Quarter 4 forecast horizon}

\begin{figure}[htb]
\centering
\includegraphics[scale=0.42]{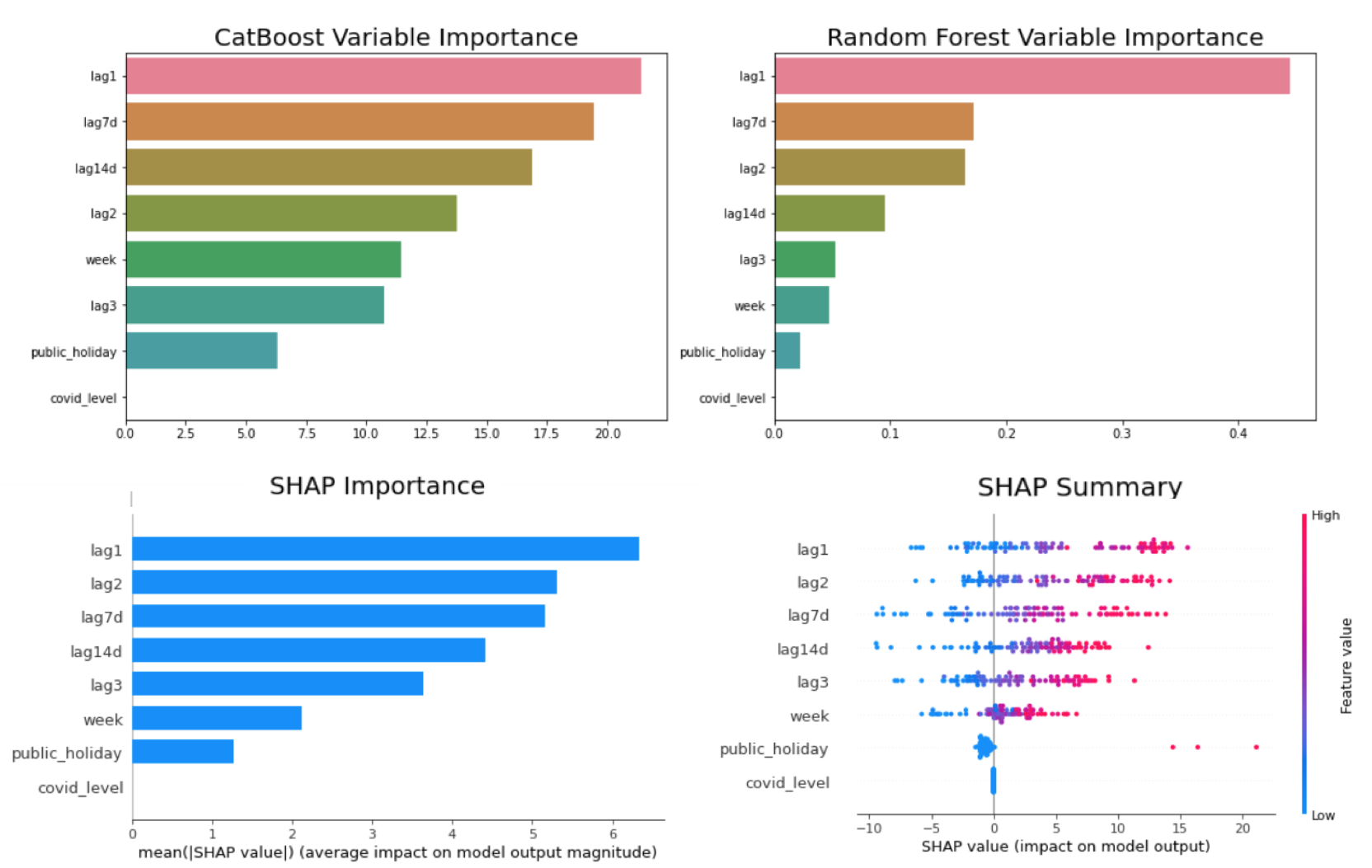}
\caption{Variable Importance Plots Q4 2019}
\label{figure:imp_2019}
\end{figure}

The variable importance plots and SHAP Summary plots in Figure \ref{figure:imp_2019} generated from the forecast in Quarter 4 of 2019 shows similar results to that of Quarter 2 2018 for the top few features, with the demand 1 year ago being the most important. Demand seven days prior is the next important feature in the variable importance plots however the SHAP diagrams place demand 2 years prior as the next important. The consistency between the two different forecasting horizons is representative of feature importance plots from other forecasting horizons drawn from stable non-pandemic scenarios (2017-2019). We can therefore conclude that during typical operating conditions, the most impactful features are the demand from the previous year for a given day which offer stability to the forecasts, together with the demand from 7 and 14 days prior, which capture fluctuations and responsiveness to dynamic conditions as they evolve - such as seasonal flu or other large outbreaks of respiratory illnesses which have a stochastic component to them.  

The graphs for this period do indicate that the week number has more of an influence than in Quarter 2 of 2018. As seen in the SHAP Summary plot, there are three instances flagged as public holidays which are driving a higher than average forecasted demand. These holidays are Labour Day, Christmas Day and Boxing Day.  

\subsubsection{2021 Quarter 1 forecast horizon}
The data for 2021 Quarter 1 embodies patterns that deviate from the historical record due to the pandemic conditions and imposed lockdowns which can be seen in Figure \ref{figure:ts}. Therefore, this forecasting horizon represents an instructive illustration of the role that different features play in adjusting the forecasts of this period and the overall adaptability of the model to disruptions of typical patterns. 

\begin{figure}[htb]
\centering
\includegraphics[scale=0.42]{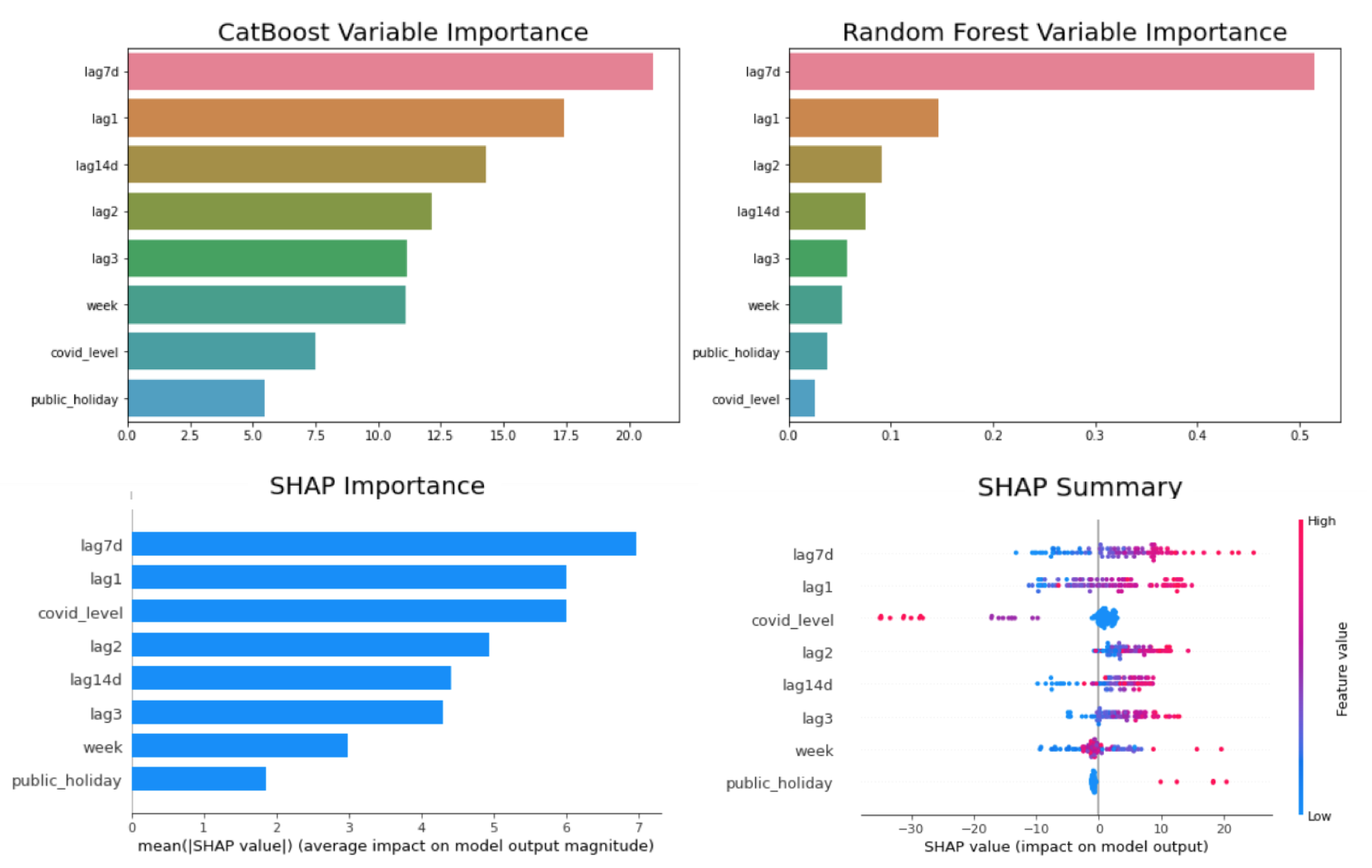}
\caption{Variable Importance Plots Q1 2021}
\label{figure:imp_2021}
\end{figure}

The variable importance plots in Figure \ref{figure:imp_2021} depict feature importances for Quarter 1 of 2021. We can observe clearly from the plots that recency expressed as lag values from the previous week have become the most important drivers of forecasted values rather than the more erratic COVID-19 2020 demand values from one year prior. This is in contrast to previous forecast periods where the most important feature was the demand from one year ago, thus indicating that the expectations of stability were held by the prior models. We can also see especially in the case of the SHAP importance graph that the COVID-19 alert level has become a significant contributor to the final predictions too for the Q1 2021 period, unlike in the more stable forecast vintages.  
From the SHAP Summary plot it can be seen how higher COVID-19 level values have a larger impact on reducing the predicted demand, with low values having little impact.  This behaviour is expected, as on average COVID-19 Alert Level 4 days witnessed a significant reduction in patients compared to COVID-19 Alert Level 1 days.  \par

\begin{figure}[htb]
\centering
\includegraphics[scale=0.75]{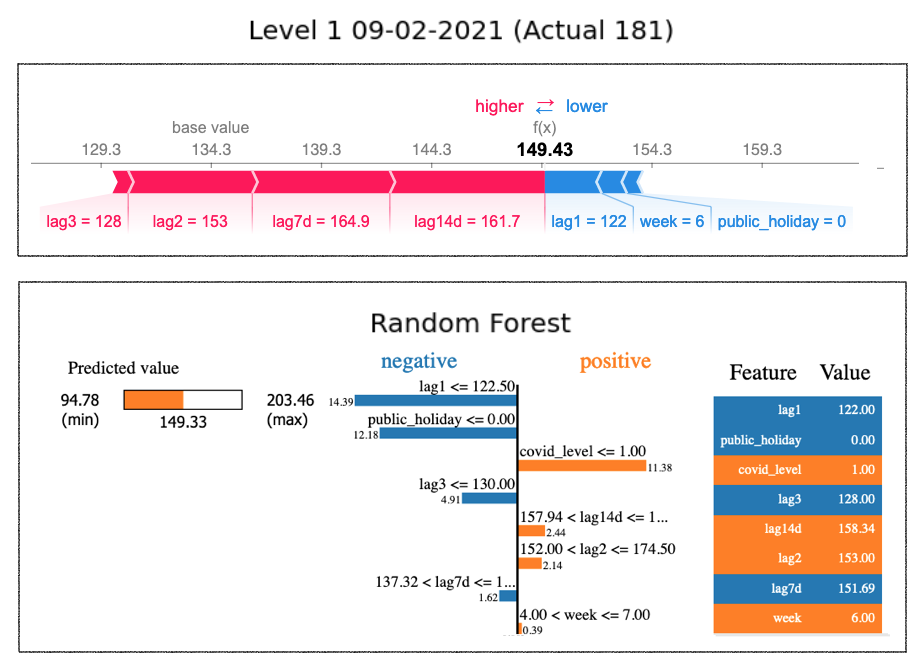}
\caption{COVID-19 Level 1 2021}
\label{figure:level1_2021}
\end{figure}

\subsubsection{COVID-19 Alert Level model behaviour}

\begin{figure}[htb]
\centering
\includegraphics[scale=0.75]{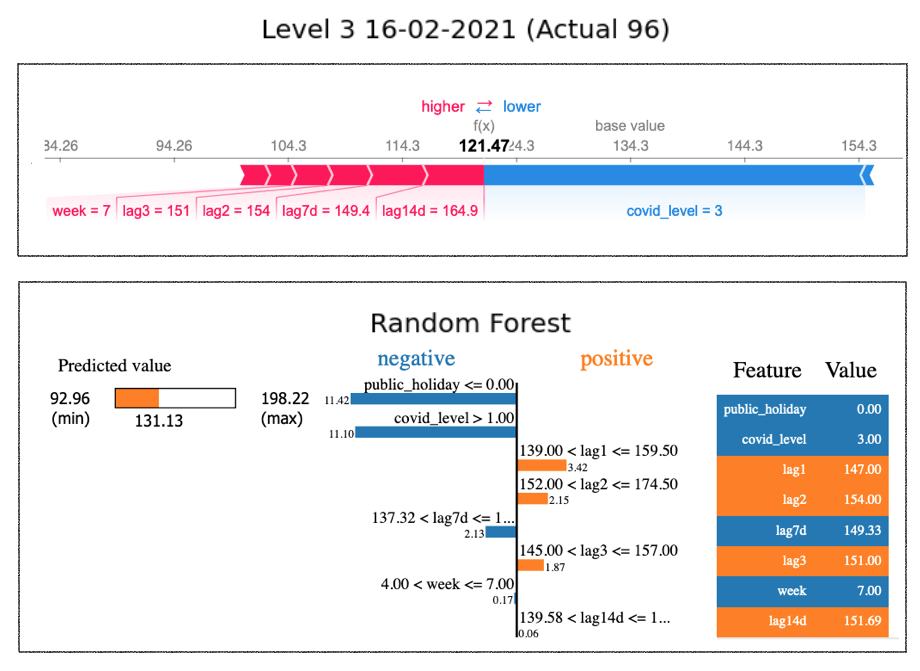}
\caption{COVID-19 Level 3 2021}
\label{figure:level3_2021}
\end{figure}

We demonstrate the effects of the COVID-19 Alert Level values in greater detail through the model explainability perspective which exposes precisely how the model reasons about its forecasts. We demonstrate the forecasting behaviour of the models on two different days, one forecast for a day from a period representing COVID-19 Alert Level 1 and another from a day during the COVID-19 Level 3 conditions. Figures \ref{figure:level1_2021} and  \ref{figure:level3_2021} depict the differences between predictions for Alert Level 1 and Alert Level 3 COVID-19 days.  

On the Alert Level 1 day, Tuesday 9 February 2021, the demand from seven and 14 days prior were higher than average and hence had a positive effect on predicting demand. However the demand from one year prior was significantly lower than the average for this feature giving a negative SHAP value, so it diminished the forecast slightly.  Exactly one week later, Auckland was in Alert Level 3. Figure \ref{figure:level3_2021} shows that the positive effects from the lag values were small compared to the large negative effect that the COVID-19 level indicator played resulting in the forecasted demand being significantly reduced.

\subsubsection{Easter Monday model behaviour}
\begin{figure}[htb]
\centering
\includegraphics[scale=0.72]{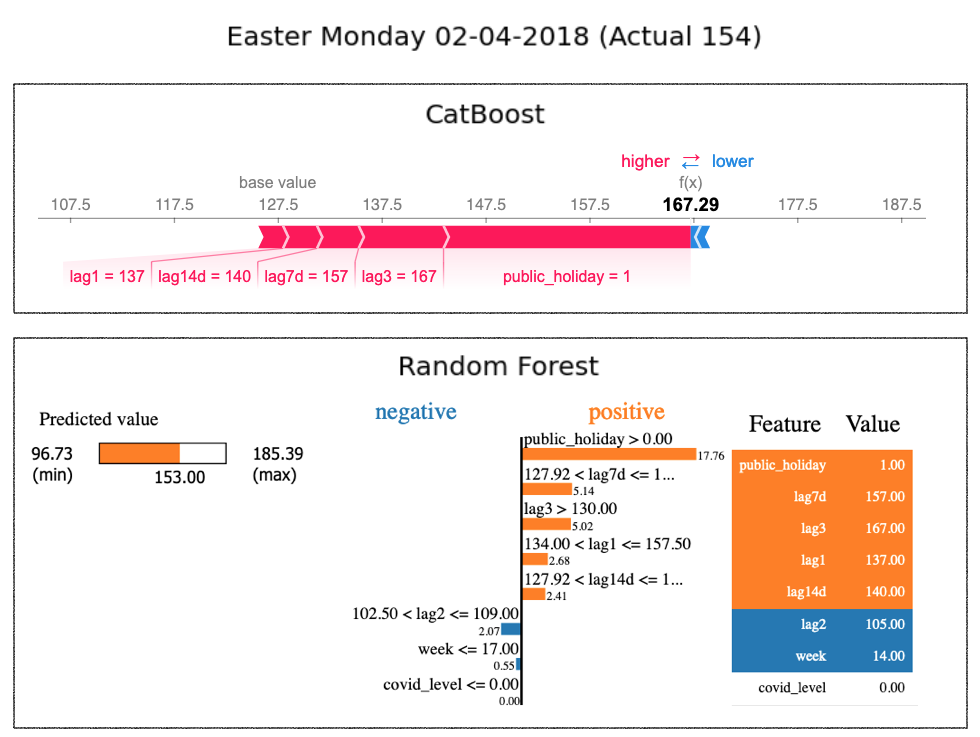}
\caption{Easter Monday 2018}
\label{figure:easter_2018}
\end{figure}

We now highlight the effectiveness of the public holiday feature we created to impact forecast results. We use Easter Monday (02-04-2018) to illustrate the model mechanics on this scenario. The actual observed demand on this day was 154 patients.  CatBoost predicted 167 and Random Forest 153. Figure \ref{figure:easter_2018} shows the SHAP interpretation of the CatBoost model on this day, and the LIME interpretation of the Random Forest model. \par

The SHAP force plot depicts for a single given day the impact each feature has on the resulting prediction. The base value (127.5) is the average predicted demand over the entire dataset.  Features having a positive impact increasing the forecasted demand are shown by a red arrow pointing to the right and pushing the forecast in this direction away from the baseline value. Features having a negative impact reducing the prediction are shown by a blue arrow pointing to the left.  The length of the arrow depicts the magnitude of the feature's contribution.\par

Flagging the day as a public holiday contributes the most to increasing the prediction from its expected base value of 127.5  to the final predicted value of 167.29.  This can be seen from the high SHAP value for Easter Monday in the SHAP Summary plot in Figure \ref{figure:imp_2018} for the public holiday feature.

The mean value for both the demand one year prior and seven days prior is 135.  On Easter Monday the demand from one year prior of 137 is around the base value and increases the prediction only slightly, with the demand seven days prior having a higher feature value of 157 leading to a higher SHAP value and a larger contribution to increasing demand.  Despite the demand three years prior only ranking as the fourth important variable, on Easter Monday this value is 167 which is significantly higher than the average of 115, resulting in a higher SHAP value and hence it has a large influence on increasing the prediction.\par

The two features in blue negatively impacting demand slightly are the demand from two years prior and the week number.  The mean value of the demand two years prior is 120 and on Easter Monday this value is only 105 with a negative SHAP value hence the prediction is reduced slightly.  With Easter Monday falling in week 14 this impacts negatively on the prediction.

The Random Forest model provides a more accurate prediction for this day. The LIME interpretation plot for this model shows the impact each feature has on the prediction, with the middle section of the diagram showing the most important feature at the top and least important at the bottom.  Features having a positive impact will increase demand and are shown in orange, with those features negatively impacting demand in blue.  The bar line on the left side of the diagram shows where the prediction sits within the range of prediction values, with Easter Monday predicting slightly higher than average demand.  The right side of the diagram lists each feature and their associated values for this day.  As with the CatBoost model, flagging the day as a public holiday has the largest impact on increasing the prediction.  The demand seven days prior has slightly more impact on increasing the prediction than the demand three years prior. The demand one year prior and a fortnight prior have a smaller impact on increasing the forecast, with the demand two years prior and the week number decreasing the demand slightly.

The results from the figure clearly indicate the utility and effectiveness of the public holiday feature in its ability to adjust forecasts accordingly and ultimately produce more accurate forecasts. 

\subsubsection{School Holidays model behaviour}

\begin{figure}[htb]
\centering
\includegraphics[scale=0.75]{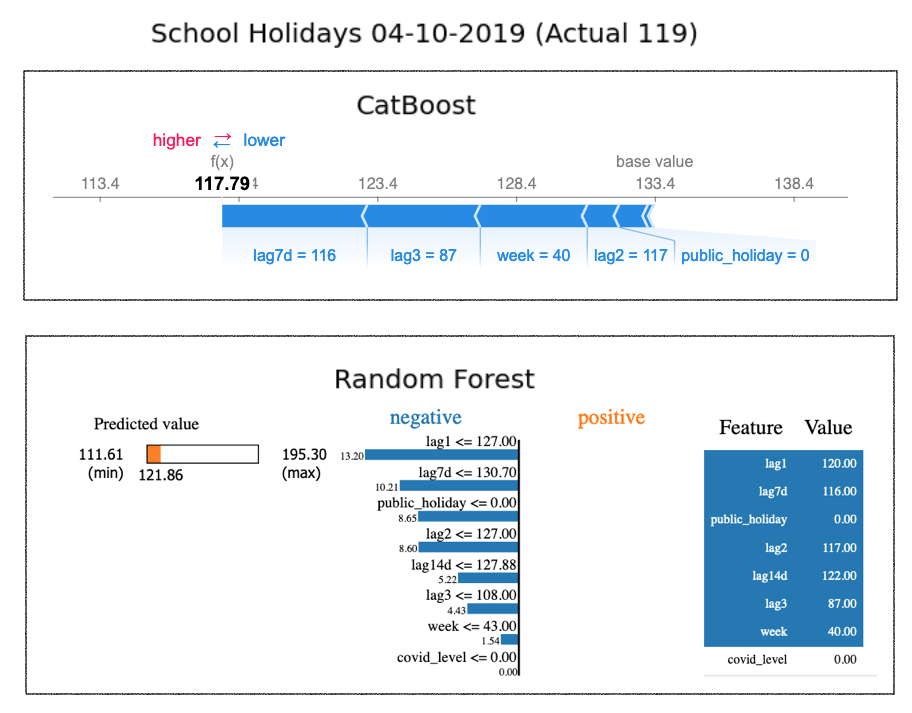}
\caption{School Holidays 2019}
\label{figure:school_holidays_2019}
\end{figure}

Finally, we illustrate the model behaviour on days falling during school holidays. We know that school holidays have an effect on patient presentation volumes at both clinics. The effect tends to decrease demand. We highlight an example of typical model behaviour on a school holiday drawn from Friday 04-10-2019. The model behaviour for this day can be seen in Figure \ref{figure:school_holidays_2019}. 

Figure \ref{figure:school_holidays_2019} shows that the predicted demand is impacted from the lower demand in all lag values which fall well below the mean value for each feature giving negative SHAP values for the CatBoost model and negative effects in the Random Forest model.  The LIME bar line shows that the forecast lies at the lower end of the range of predictions. These lower values are expected as school holidays fell during the same time the previous year, also with a lower demand seen.  This demonstrates the lag values used in the models naturally incorporate adjustments to lower the predicted demand during school holidays without the need for engineering extra model features. 

When relying on school holidays falling during the same time the previous year, problems may be evident when school holidays do not coincide. This sometimes occurs at the end of school Term 1.  This may impact the Random Forest model more than CatBoost as it places heavy importance on the demand one year ago. Actual observed demand on this day was 119 patients, with both models providing close estimates of 118 for CatBoost and 122 for Random Forest.

\section{Discussion}
In response to RQ1, our research conclusively indicates that it is indeed possible to leverage machine learning algorithms in order to generate patient demand forecasts at the UCCs which improve on existing in-house strategies. It has been demonstrated that prior to the COVID-19 pandemic, demand can be predicted to a reasonably acceptable level of accuracy. Forecasts made on data prior to COVID-19 pandemic conditions indicated that the improvement of the machine learning models over the in-house benchmark strategies ranged between 23\%-27\% across both clinics.   

Following the outbreak of the COVID-19 pandemic in New Zealand, and the unpredictability of constraints placed on residents, resulting model errors were exacerbated by the challenges of modelling under such erratic circumstances. However, the generated forecasts still displayed a significant improvement over the in-house models as well as the competing benchmarking approaches to estimation explored in this study.  \par

The requirement of the models to generate reliable forecasts three months ahead is demanding since there are many unaccounted factors which can occur and heavily influence eventual demand. Naturally, these factors cannot be captured at the time of forecasting, but all things being equal, the best performing models exhibited significant reliability even as the forecasts extended out towards the most distant forecast horizon. The smaller variation shown in the first week from using observed rather than forecasted 7 and 14 day lag values shows the advantages accompanying proximal forecasts in terms of higher accuracy. As the forecast horizon increased, the following weeks produced remarkably stable variations in residuals up until week 13 where the forecasts started to exhibit some signs of a larger variation.  

A broad range of machine learning algorithms were explored in this study. Literature testifies to the advantages of using ensemble-based methods over single models. The theory of ensemble-based machine learning states that aggregations from diverse predictive models will on average outperform single models across numerous datasets. This theory has repeatedly been upheld and the same has been demonstrated in this study. In answering RQ2, we find that the Voting algorithm has on average demonstrated better generalisability properties over the other algorithms for this particular domain and these specific datasets. The base estimators which constitute the Voting model were chosen both on the merits of their individual performances but also on the basis that they were sufficiently diverse from one another, which is a cornerstone principle of effective ensemble-based modelling.

Not only is producing reliable forecasts three months ahead demanding due to numerous unforeseen factors which can affect long-term demand, but it is also difficult in respect to the limitations it places on the types of features that can be used for making predictions. It is a fact that numerous proxy variables capturing data on current weather, temperature, traffic as well as internet search terms on  medical ailments and even flu-tracker data hold potential for accurately estimating very short term patient demand in UCCs. However, while these proxy variables for predicting demand are often indicative of immanent patient presentations, their ability to be used for predicting patient presentations one month, two months or three months ahead is extremely tenuous. Therefore, in practice there is a fairly limited number of variables which can be used with machine learning this problem domain.    

Given this constraint, models generated in this analysis predominately used lag values and flags for public holidays and COVID-19 Alert Levels to provide information to the algorithms. 

The domain experts in the area of estimating patient demand from the clinics in this study indicated that  following a public holiday, demand is often accentuated as patients delay care until after the holiday.  Rigorous testing of the models was performed to determine if the addition of an extra flag to indicate days following a public holiday was advantageous, however CatBoost and Random Forest showed little gains, with risks of over-fitting.  The Prophet model regularly showed improved accuracy and hence the upper window flag of the Prophet model was set to reflect this. With Voting selected as the optimal model, of which Prophet was a contributing algorithm, this pattern in the data is incorporated to some degree.  \citet{Batal2001} also highlight the importance of this post-holiday demand increase, which they note \citet{Holleman1996} observed too, hence they used an extra variable to represent these days.\par

Therefore, in addressing RQ3, we found that during stable (non-COVID-19) periods the most useful features were lag values of actual demand from one year prior, followed by lag values from one week prior. However, this changed when forecasts were made for COVID-19 periods which were accompanied by various lockdown mandates. During these periods, we found that lag values of one week prior became the most impactful, followed by the lag values of one year before as well as the variable indicating the current COVID-19 Alert Level. It is interesting to note that hospital EDs experienced unprecedented declines in patient volumes during the pandemic crisis \cite{hollander2021availablists}. Given the findings from this study, forecasting models specifically for EDs could in future therefore also benefit from integrating variables which capture COVID-19 or other pandemic alert levels in order to produce more accurate estimates. 

\section{Conclusion}
The ability to accurately forecast patient demand in Urgent Care Clinics (UCCs) and Emergency Departments (EDs) is becoming increasingly expected so that an efficient allocation of human resources can be realised in order to  prevent congestion, and deliver a consistently high-quality of medical care. 

Forecasting patient arrivals is however a challenging undertaking with many latent factors ultimately affecting patient presentation volumes. The problem becomes even more difficult as the forecast horizons are extended and required to generate estimates of patient demand further out in the future.

Up to now, research efforts to develop forecasting models for this problem domain have predominantly involved traditional statistical methods. In this study we have explored the feasibility of machine learning models to generate accurate forecasts of up to three months ahead for two large UCC clinics in Auckland, New Zealand. Our work demonstrated that the machine learning models were significantly more effective at forecasting patient demand than existing in-house methods. The study determined that ensemble-based methods produced the most accurate results on average.

This paper also provided a rigorous analysis into understanding the behaviour of the machine learning models, looking specifically at what features were most impactful, as well as how the values for these features affected the actual forecasts. The study also took into account the COVID-19 pandemic conditions and the affects these conditions had on patient presentations and the ability of the machine learning models to adapt to the ensuing volatility. We found that the most effective features during the pre-pandemic periods were lagging variables from the same periods in previous years, while lagging values from more recent periods which were more effective at capturing evolving patterns together, with data indicating COVID-19 Alert Levels were the most useful during the pandemic periods.

\urlstyle{same}

\bibliography{main}

\end{document}